\documentclass{article}

\usepackage{microtype}
\usepackage{graphicx}
\usepackage{subcaption}
\usepackage{booktabs} 
\usepackage{multirow}
\usepackage{soul}

\usepackage{hyperref}

\usepackage{xurl}

 \usepackage[preprint]{icml2026}

\usepackage{amsmath}
\usepackage{amssymb}
\usepackage{mathtools}
\usepackage{amsthm}
\usepackage{pifont}

\usepackage[capitalize,noabbrev]{cleveref}
\theoremstyle{plain}

\theoremstyle{definition}

\theoremstyle{remark}

\usepackage[textsize=tiny]{todonotes}
\usepackage{comment} 

\icmltitlerunning{FAMOSE: A ReAct Approach to Automated Feature Discovery}

\begin{document}

\twocolumn[
  
  \icmltitle{FAMOSE: A ReAct Approach to Automated Feature Discovery}



  \icmlsetsymbol{equal}{*}

  \begin{icmlauthorlist}
    \icmlauthor{Keith Burghardt}{1}
    \icmlauthor{Jienan Liu}{1}
    \icmlauthor{Sadman Sakib}{1}
    \icmlauthor{Yuning Hao}{1}
    \icmlauthor{Bo Li}{1}
  \end{icmlauthorlist}
  \icmlaffiliation{1}{Amazon.com, Inc.,  410 Terry Ave N, Seattle, WA 98109, United States}

  \icmlcorrespondingauthor{Keith Burghardt}{kaburg@amazon.com}

  \icmlkeywords{Feature Engineering, AI Agent, LLM, Supervised Learning}

  \vskip 0.3in
]



\printAffiliationsAndNotice{}  

\begin{abstract}
Feature engineering remains a critical yet challenging bottleneck in machine learning, particularly for tabular data, as identifying optimal features from an exponentially large feature space traditionally demands substantial domain expertise. To address this challenge, we introduce \textbf{FAMOSE} (\textbf{F}eature \textbf{A}ug\textbf{M}entation and \textbf{O}ptimal \textbf{S}election ag\textbf{E}nt), a novel framework that leverages the ReAct paradigm to autonomously explore, generate, and refine features while integrating feature selection and evaluation tools within an agent architecture. To our knowledge, FAMOSE represents the first application of an agentic ReAct framework to automated feature engineering, especially for both regression and classification tasks. Extensive experiments demonstrate that FAMOSE is at or near the state-of-the-art on classification tasks (especially tasks with more than 10K instances, where ROC-AUC increases 0.23\% on average), and achieves the state-of-the-art for regression tasks by reducing RMSE by 2.0\% on average, while remaining more robust to errors than other algorithms. We hypothesize that FAMOSE's strong performance is because ReAct allows the LLM context window to record (via iterative feature discovery and evaluation steps) what features did or did not work. This is similar to a few-shot prompt and guides the LLM to invent better, more innovative features. Our work offers evidence that AI agents are remarkably effective in solving problems that require highly inventive solutions, such as feature engineering. 
\end{abstract}

\section{Introduction}


Despite the simplicity of tabular models, their performance is often difficult to optimize. Namely, while carefully engineered features are often the most important component to model performance \cite{Heaton2016,tschalzev2024data}, their optimization remains a difficult and time-consuming bottleneck. This is because identifying those valuable features is extremely difficult: the space of potential new features (transformations or combinations of existing attributes) is enormous, and only a small fraction will ultimately prove useful. As a result, designing features has traditionally relied on human intuition and domain knowledge, and it continues to consume a significant portion of a data scientist’s effort \cite{bengio2013representation}. 

\begin{figure}
    \centering
    \includegraphics[width=0.8\linewidth]{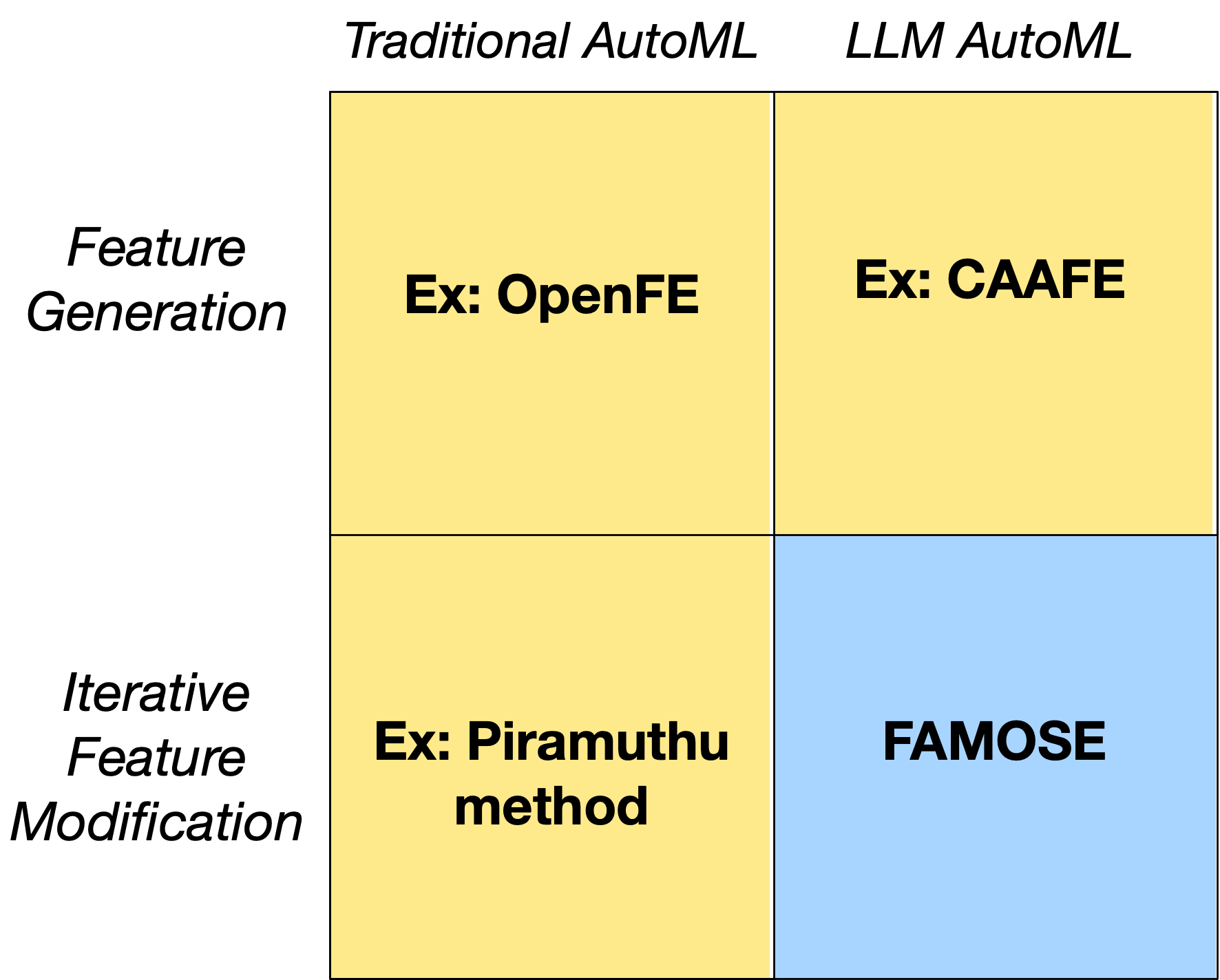}
    \caption{Types of AutoML. In traditional AutoML, feature engineering would include feature discovery, such as within OpenFE \cite{zhang2023openfe}, or iterative feature modification, such as the method of Piramuthu et al. \cite{piramuthu2009iterative}. Although LLM methods, such as CAAFE, often excel at feature generation, iterative feature modification has not been explored as often before now. FAMOSE offers a way to learn through trial-and-error which features work and which ones do not until a better feature is developed.}
    \label{fig:FAMOSEDifference}
\end{figure}

To reduce the manual burden, a variety of automated feature engineering techniques have been proposed, whether algorithmic methods, such as OpenFE \cite{zhang2023openfe}, or LLM-based methods, such as CAAFE \cite{Hollmann2023}. However, these methods struggle due to the combinatorically large space of features to evaluate \cite{zhang2023openfe,horn2019autofeat}. While LLM-based methods may offer a route to search this space more efficiently, these methods do not iteratively improve on their prior features by learning from their mistakes. This can be illustrated in Fig.~\ref{fig:FAMOSEDifference}.

To address these limitations, we introduce \textbf{FAMOSE}: \textbf{F}eature \textbf{A}ug\textbf{M}entation and \textbf{O}ptimal \textbf{S}election agEnt, a framework that utilizes ReAct agents for autonomous feature engineering. The agent iteratively proposes, evaluates, and refines features, rather than generating a static list in a single pass. {FAMOSE} therefore simulates a data scientist who iteratively hypothesizes, tests, and refines features to improve predictions.

Inspired by the Reasoning and Acting paradigm, the agent can reason about the task, execute code, and adapt its strategy based on validation performance, enabling a feedback-driven search process similar to how human practitioners experiment with features. Unlike one-shot or template-based approaches, the agent can directly interact with the dataset to compute statistics, inspect distributions, and test hypotheses about feature utility. As a result, feature proposals are grounded in the actual data rather than relying solely on prior knowledge. The iterative loop also allows the agent to detect and correct invalid or unhelpful features, reducing hallucinations and improving robustness. By learning from empirical outcomes, the agent can more effectively navigate the feature space and focus on transformations that are informative in context. Moreover, during the ReAct process, the LLM explains the reason it chose this feature, improving feature interpretability. At the end of each round, the best feature that improves model performance (if one is found) is saved. At the next round, we re-start the ReAct agent to create yet more features. Each new feature's performance is then evaluated conditional on the original features and all the saved features from prior rounds. Once performance stops increasing at the feature discovery step, FAMOSE then applies a minimal-redundancy maximal-relevance (mRMR) feature selection step \cite{ding2005minimum} to produce a compact final feature set. This final step contrasts with prior work that used the LLM itself for feature selection \cite{Hollmann2023}. We expect that algorithms might select features more accurately than LLMs, so mRMR is chosen as a reasonable feature selection tool. The entire process is fully automated and generalizes across data classification and regression with diverse dataset sizes and feature dimensionalities.

In summary, our contributions are as follows:
\begin{itemize}
    \item \textbf{ReAct-based feature discovery.} We propose a ReAct agent framework to autonomously generate, evaluate, and refine features through iterative interaction with the data.
    \item \textbf{Comprehensive benchmarking.} We evaluate our method on a diverse suite of classification and benchmark datasets, and check model robustness via evaluating its performance on multiple tabular models as well as an alternative LLM.    
    \item \textbf{State-of-the-art performance and robustness.} The proposed method matches or exceeds state-of-the-art performance across classification and regression tasks, and can create features in datasets where some autoML methods fail. This is especially true for large classification datasets (e.g., more than 10K instances, where FAMOSE's performance is substantially higher than competing methods).
    \item \textbf{Feature interpretability through reasoning.} We leverage LLM-based reasoning to provide human-interpretable explanations for why generated features are useful.
\end{itemize}

Overall, this algorithm offers a simple, scalable, and robust method for feature discovery. Moreover, while LLMs have been understood as general pattern machines \cite{mirchandani2023large}, our work offers evidence that AI agents can be remarkably effective at solving problems that need highly innovative solutions, despite the limitations of LLMs at creating diverse outputs \cite{jiang2025artificial}.

\section{Related Work}
Automated feature engineering has long been studied to improve prediction performance, where carefully engineered features have been shown to outperform gains from model selection or hyperparameter tuning alone \cite{Heaton2016}. Early work focused on pre-processing data, namely creating compact or latent representations of data, including principal component analysis (PCA) \cite{abdi2010principal}. In PCA, a new set of orthogonal features are created that represent the largest amount of variance in the data. PCA can be used to reduce the feature space into a subset of distinct features, although this only creates features that are linear functions of the original features. 

A substantial body of work, however, has investigated combinatorial and search-based feature engineering. This includes AutoFeat \cite{horn2019autofeat}, DIFER \cite{zhu2022difer}, and OpenFE \cite{zhang2023openfe}, which each create a set of new features based on transformations and operations on older features and then prune the resulting feature set. Other examples include ExploreKit \cite{Katz2016ExploreKitAF} and Deep Feature Synthesis \cite{Kanter2015DeepFS} (also cf. \cite{Heaton2016,nargesian2017learning,davis2016automated,nargesian2017learning}), which generate candidate features by enumerating transformations over existing features, then follow a heuristic or model-based evaluation for feature selection. While these approaches can produce interpretable features, they often struggle to scale efficiently with memory or time because the search space increases enormously with the number of features in the original dataset. Moreover, because the transformation rules are predefined, these methods may miss domain-specific or multi-step feature constructions that fall outside their templates. An example of multi-step feature construction, however, includes \cite{piramuthu2009iterative}, which iteratively creates new features based on features it already created, but this idea has not been as carefully explored in autoML literature.


Recent literature has begun to address the limitations of traditional autoML with LLMs. Namely, LLMs have recently been utilized for feature generation or transformation \cite{Hollmann2023,Zhang2024,han2024large,nam2024optimized,he2025fastft,abhyankar2025llm}, which expands on prior work utilizing LLMs for tabular data understanding \cite{Fang2024,wang2023unipredict,dinh2022lift,inala2024data}. A state-of-the-art LLM based feature engineering tool is CAAFE \cite{Hollmann2023}, which uses an LLM to suggest new features based on dataset metadata and task descriptions. In contrast, FeatLLM \cite{han2024large} treats feature generation as an optimization problem guided by LLM proposals. Related efforts explore evaluating or benchmarking LLM-generated features \cite{Zhang2024,bordt2024elephants} or guiding feature construction using model-based reasoning \cite{nam2024optimized}. These methods demonstrate that LLMs can propose creative and context-aware features that go beyond simple transformations. However, most existing LLM-based approaches follow a one-shot or template-driven workflow: features are generated once, evaluated externally, and not iteratively refined by the LLM itself. As a result, the model cannot learn from feedback, correct invalid feature definitions, or adapt its strategy based on empirical performance. 

Once features are generated, a large body of work has developed feature selection tools \cite{Chandrashekar2014}. Many heuristics have been explored (cf. \cite{zhang2023openfe,horn2019autofeat,Katz2016ExploreKitAF,Kanter2015DeepFS}, etc.), which we will not have space to explore in depth. However, one widely-used selection method is minimum redundancy–maximum relevance (mRMR) \cite{ding2005minimum}. This method selects a subset of features based on minimizing redundancy (e.g., minimizing correlations between features) while maximizing the relevance (e.g., finding features highly correlated with the target variable). Other methods of feature selection include LLMs \cite{Hollmann2023}. Although we expect algorithmic feature selection is more accurate than LLM-based methods, a comprehensive comparison between these methods have not been explored.

In contrast to prior approaches, our work's most substantial contribution is that {FAMOSE} uses a ReAct framework \cite{yao2022react}, in which the LLM iteratively proposes, evaluates, and refines features based on model feedback and data-driven analysis. By integrating reasoning, acting, and validation within a closed loop, this approach aims to address the scalability, robustness, and adaptability limitations observed in prior automated feature engineering methods. In addition, our work conditions the performance on new features with the performance of the model with all prior picked features, allowing us to find features that add more signal to our prediction task than all previous features. Finally, we utilize a traditional algorithm for feature selection (mRMR \cite{ding2005minimum}) rather than an LLM \cite{Hollmann2023}, which we believe allows for fast and accurate feature selection.

\section{Methodology}
In this section, we describe the {FAMOSE} methodology for feature generation, the datasets used to evaluate FAMOSE, the experimental setup, and the baseline algorithms.

\begin{algorithm}[t]
\caption{FAMOSE:}
\label{alg:craft}
\begin{algorithmic}[1]
\STATE \textbf{Input:} Dataset $D=(X,y)$, base model $\mathcal{M}$
\STATE \textbf{Output:} Selected engineered feature set $\mathcal{F}^{\star}$
\STATE Split $D$ into $K{=}5$ folds (stratified if classification)
\FOR{Each fold}
    \STATE Split training data into train and validation
    \STATE Initialize feature set $\mathcal{F} \leftarrow \emptyset$
    \STATE Define $E(Z)$, the error of a model in validation data
    \STATE Evaluate baseline score on validation set
    \FOR{$r = 1$ to $20$}
        \FOR{step$ = 1$ to $10$}
            \STATE Agent proposes a feature, $f$, using metadata and tool feedback
            \STATE Validate feature code 
            \IF{validation fails}
                \STATE Agent creates new code
            \ENDIF
            \IF{$1-E(X\cap\mathcal{F}\cap \{f\})/E(X\cap\mathcal{F})<0.01$}
                \STATE Continue
            \ENDIF
        \ENDFOR
        
        Find $f$ s.t. $E(X\cap\mathcal{F}\cap \{f\})$ is maximized
        \IF{$E(X\cap\mathcal{F}\cap f) > 0$}
            \STATE $\mathcal{F}\leftarrow \mathcal{F}\cap \{f\}$
        \ENDIF
    \ENDFOR
    \STATE Apply mRMR to select final features $\mathcal{F}^{\star}$
    \STATE Train $\mathcal{M}$ using $\mathcal{F}^{\star}$ and evaluate on test fold
\ENDFOR
\end{algorithmic}
\end{algorithm}

\subsection{FAMOSE}
\begin{figure}
    \centering
    \includegraphics[width=1.0\linewidth]{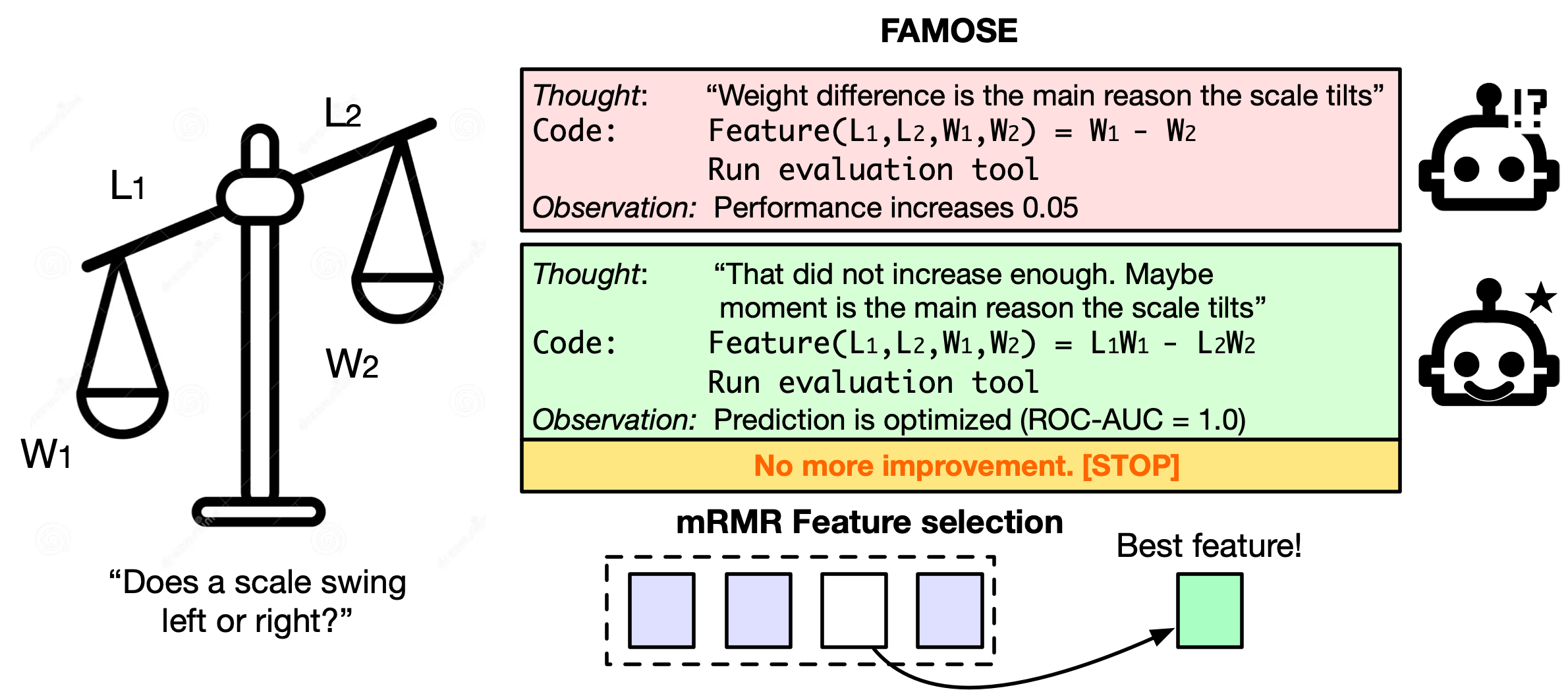}
    \caption{An illustration of {FAMOSE} applied to the \texttt{balance-scale} task \cite{balance_scale_12}. {FAMOSE} first invents features and observes their performance, in this case using the difference in arm weights to see whether a scale is balanced, tilting right, or tilting left. If performance is insufficient, {FAMOSE} then thinks about how to create better features, ultimately discovering the feature, moment, that exactly predicts whether a scale is balanced. Because the performance cannot be improved further in this task, the feature discovery step eventually stops. The feature selection step then selects this feature using mRMR \cite{ding2005minimum} and removes the four extraneous features in the task dataset: $W_1$, $W_2$, $L_1$, and $L_2$, thus instead of an error-prone model with four features, we discover a perfect predictor with only one feature.}
    \label{fig:example}
\end{figure}

We outline {FAMOSE} in Algorithm~\ref{alg:craft}, with an example of {FAMOSE} in Fig.~\ref{fig:example}. {FAMOSE} at it's core is Smolagents \cite{smolagents} with additional tools and a post-processing evaluation algorithm to ensure that the best features are selected. Smolagents is an opensource AI agent that can think, act, and observe, but can hallucinate without appropriate tools. We removed its core tools except for the Python code compiler, and created a metadata generator tool that allows the agent to determine data file column names (features) and whether the column values represent numerical, datetime, or categorical data. This allows the agent to better understand features without manually incorporating those details into the prompt. We also provided it with a feature evaluation tool, in which code generated by an LLM is used to generate a feature column. The model performance (ROC-AUC or RMSE) in held-out data containing the candidate feature, the original features, and all of the best features chosen so far is compared against a model without the candidate feature. If the change in performance is positive, the feature improves model performance. We negate RMSE so that a positive change always represents a better model. The tool is highly robust with the help of error correction and regex to extract all the feature names in the feature code; if the feature names include hallucinated features, or if the code cannot run, then the tool forces the agent to correct its mistakes, including the replacement of hallucinated features with correct ones. Data provided to the feature generation code also has the target variable removed to avoid data leakage.

We provide the following prompt for the agent (see details of prompt and agent output in the Appendix):
We first specify a role (data scientist) and the relevant files and feature descriptions (see example feature descriptions in Appendix Tables~\ref{tab:feature_desc1} \& \ref{tab:feature_desc1}). We then ask the agent to
\begin{enumerate}
    \item Use all of these insights to create a large set of new features. It can use any mathematical operations or transforms.
    \item Explain why this feature is useful
    \item Check the performance of each feature with a feature evaluation tool
    \item Set a goal to improve feature performance by 1\%, and do not stop until this is achieved.
    \item Return the best feature.
\end{enumerate}
The agent will then autonomously use the file meta-data tool, create Python code, and evaluate features with the appropriate tools. If the agent discovers that a feature did not improve performance by at least 1\%, it will iteratively improve on and evaluate new features for up to 10 agent steps until it is satisfied (top right of Fig.~\ref{fig:example}). At this point, the agent will stop, and a feature evaluator algorithm will then evaluate all the features created by the agent and pick the best feature that improves model performance (if any are found). This post-agent step addresses situations in which the LLM hallucinates model performance in order to satisfy the 1\% model performance improvement goal. Finally the agent repeats all these steps for up to 20 rounds or until the performance has not improved after 6 rounds (this stops the algorithm early to save time). Adding features, however, could contribute to overfitting a model, so we use mRMR \cite{gauch2003scientific} to select the best features, as shown in the bottom right of Fig.~\ref{fig:example}. This contrasts with other SOTA feature engineering methods \cite{Hollmann2023,perez2025llm}, in which the LLM itself decides what features to remove. We specifically use 5-fold cross-validation within the training dataset to find the optimal number of features to select (based on ROC-AUC or RMSE), then apply mRMR to the training dataset to select the specific features.

\subsection{Data}
We evaluate our approach on a diverse collection of public tabular datasets, including 20 classification tasks and 7 regression tasks, spanning over three orders of magnitude in size (all datasets are outline in Appendix Table~\ref{tab:data_desc}). These dataset tasks cover binary and multi-class classification (e.g. income prediction, medical diagnosis, game outcomes) as well as regression problems (e.g. house prices, insurance costs). We specifically chose these datasets from a prior paper \cite{abhyankar2025llm}, but added additional tasks from Kaggle that contain over 10K instances to ensure that ROC-AUC differences were more statistically significant. Most datasets analyzed are commonly used as benchmarks, and a smaller subset of these have been used in many prior papers \cite{horn2019autofeat,Hollmann2023,abhyankar2025llm,han2024large}.
We borrow feature descriptions and task descriptions (i.e., what the model aims to predict) from \cite{abhyankar2025llm}, although we correct feature names for \texttt{vehicle} \cite{vehicle}, 
\texttt{balance-scale} \cite{balance_scale_12}, and \texttt{cmc} \cite{contraceptive_method_choice_30}
whose feature names were mis-written in the Github code for \cite{abhyankar2025llm}. For example, column names were lower-case when they should be capitalized, or contained spaces when they should contain underscores (``\_''). For datasets not mentioned in \cite{abhyankar2025llm}, and for \texttt{bank\_marketing}
\cite{bank_marketing}, which also contained errors in feature descriptions, we instead use feature descriptions from each dataset's respective webpage (e.g., Kaggle; all dataset links are referenced in Appendix Table~\ref{tab:data_desc}). Table~\ref{tab:data_stats} summarizes the key characteristics of the datasets, including the number of instances, number of features, (for classification tasks) class imbalance (fraction of data in the smallest class), and (for regression tasks) the target variable variance. We also list in Table~\ref{tab:data_desc} a brief description of each prediction task.

For training/validation and testing, we use \texttt{scikit-learn}'s K-Folds (if regression) or StratifiedKFold (if classification) with K$=$5, and use a random seed of 42 \cite{scikit-learn}. We further stratify training data into training and validation using K$=$5 where validation is used to select {FAMOSE} features or to determine the number of features to select for mRMR. Test folds are only used for the final performance evaluation. 

\subsection{LLMs and Resources}
For LLM-based feature engineering, we use AWS Bedrock's \texttt{Sonnet 3.5 V2} LLM and \texttt{Deepseek-R1}, as they have reasonable speed, performance, and cost. While the main text describes results using \texttt{Sonnet 3.5 V2}, we also report \texttt{Deepseek-R1} in the Appendix and find it creates similar quality features. Following prior work \cite{abhyankar2025llm}, we fix the LLM's temperature to be 0.8. Fine-tuning such hyperparameters will be important research in the future. Analysis was made with AWS Sagemaker \texttt{ml.r6i.32xlarge} and \texttt{ml.m7i.48xlarge} instances for most data, but we use a GPU instance, \texttt{ml.g5.48xlarge}, to train Autogluon \cite{erickson2020autogluon} (Cf. \url{https://instances.vantage.sh/} for a summary of instance specifications). We do not expect variations in instances will have a significant impact on model performance. Because the resources needed to run LLM inference are unknown, and could affect LLM speed, we do not compare algorithm inference times, especially the speed of LLM methods against classical methods. We would expect that if classical methods were given the same resources (perhaps 10s of GPUs) as an LLM like \texttt{Sonnet 3.5 V2}, we should expect far different speeds. That said, because we have limited time to analyze data, we do not include performance for classical algorithms that created errors or take more than 10 hours for all five folds to finish within a \texttt{ml.m7i.48xlarge} instance. LLM-based methods, meanwhile, typically finish a few minutes, and therefore do not have such restrictions. Even the longest time for FAMOSE to complete all five folds of a task is roughly 6 hours (\texttt{covtype}, which contains 580K instances and 55 features). 

\subsection{Prediction Models}
We compare different feature engineering approaches using XGBoost, as this is a common high-quality tabular model baseline. While XGBoost performance could vary based on hyperparameters, we use default parameters and set the seed to 42 for consistency. 
We also apply the XGBoost-derived features to  Random Forest \cite{breiman2001random} and Autogluon in order to determine the robustness of these features across models. Other models can also be tested on FAMOSE; we believe the models we choose, however, represent exemplars of state-of-the-art tabular models: boosted trees (XGBoost), ensembles of trees (Random Forest), and weighted ensembles of deep-learning, boosted, and foundation models (Autogluon). 

\subsection{Feature Engineering Baselines}
We compare our methods against two classical methods, AutoFeat \cite{horn2019autofeat}, and OpenFE \cite{zhang2023openfe}, as well as SOTA LLM-based methods, CAAFE \cite{Hollmann2023} and FeatLLM \cite{han2024large}. For LLM-based methods, we replace a call to the Open AI API with a call to AWS Bedrock (a change of just a few lines) in order to easily switch between various candidate LLMs we aimed to use, but all other code was kept the same. 

We do not report comparisons against another recent baseline, LLM-FE \cite{abhyankar2025llm}, as their results are hard to validate: metadata was miswritten in their code, as mentioned above, and details of model implementations, including model hyperparameters, are missing. Finally, they report results, such as normalized RMSE, but do not state how they normalized RMSE (e.g., normalize by the mean or range, within the test set or training set, etc.), making the results they report unreproducible. Smaller issues include the category pre-processing step in their code converts categories into float numbers, so that the model would interpret the categories ``dog'', ``cat', and ``car'' as 0, 1, 2, where 2 is greater than 0 (we believe this is not an adequate way to represent non-cardinal categorical variables). Due to challenges reproducing their results, and the incorrect pre-processing step they use on data, we cannot be certain we replicated their work, so we leave out this model (that said, our best guess at reproducing their results shows significantly worse performance than our method or CAAFE). 

In some datasets, we find column names contain unusual characters (e.g., ``)'' or 
``$<$''), which would confuse classical methods (AutoFeat or OpenFE). AutoFeat and OpenFE read and write feather-formatted files, which cannot natively parse those characters. For the classical feature engineering methods, we replaced column names,
``('' or ``)'' with ``\_'', ``['' (``]'') with ``left(right)bracket'' and ``$<$''($>$) with ``less(greater)than''. Despite results reported in \cite{abhyankar2025llm}, the simple change in the column names dramatically improved the robustness of each of the classical methods, and the only errors were due to out-of-memory errors (e.g., creating and then selecting among millions of features would require TBs of memory), or more rarely out-of-time errors.

\section{Results}
In Table~\ref{tab:Comparison_class}, we show classifier performance: the mean ROC-AUC after applying FAMOSE, and 4 competing SOTA methods: AutoFeat \cite{horn2019autofeat} and OpenFE \cite{zhang2023openfe} (both classical methods), and FeatLLM \cite{han2024large} and CAAFE \cite{Hollmann2023} (both LLM-based methods). For any multi-class tasks, such as \texttt{balance-scale}, we calculate an unweighted average of one-versus-one ROC-AUC. 
We find that FAMOSE is statistically insignificantly different from the SOTA methods, and consistently find AutoFeat and OpenFE fail to handle particularly complex datasets or data with more than $\sim$10K rows. This is because the number of features these models analyze scale greatly with the number of columns, which in turn creates, e.g., TBs of memory for moderately large datasets. Unfortunately, this means classical methods fail for many realistic datasets. FeatLLM, meanwhile, fails or performs poorly for different reasons, often due to issues including creating rules for multiple classes, or because the LLM output will end without completing the task. CAAFE and FAMOSE, meanwhile can work across all tasks. While we find OpenFE is strongest for tasks with fewer instances, {FAMOSE} pulls away for tasks with more than 10K instances, with a \textbf{mean improvement of 0.23\%}. We chose 10K as a reasonable compromise between size and number of tasks (7 out of 21).  We compare features developed by CAAFE and {FAMOSE} in the Appendix, where we find that {FAMOSE} creates many more diverse functions.

FAMOSE also performs strongly for regression tasks, where it is better than the  competing methods, AutoFeat and OpenFE (LLM methods we evaluated do not create features for regression tasks). Namely, we find a \textbf{2.0\% reduction in RMSE on average}, as shown in Table~\ref{tab:Comparison_reg} (Wilcoxon signed rank test p-value$=0.07$ between {FAMOSE} and the baseline). Importantly, {FAMOSE} again performs well in both  small and large tasks, while classical methods often fail to evaluate larger tasks (success rate is $<90\%$). Interestingly, OpenFE performs especially poorly for the \texttt{bike} task, although we are unsure as of yet why its performance is consistently worse across all folds.

In addition, however to model performance, we see in Appendix A.2 that FAMOSE's output achieves human interpretability. For example, when applying {FAMOSE} to the \texttt{balance-scale} task, we see, much like~\ref{fig:example}, that the agent can test and discover moment difference is the only feature needed to capture whether a scale is balanced or tilts. 

We apply an alternative LLM, Deepseek-R1 \cite{guo2025deepseek}, to {FAMOSE} and discover performance is similar (see Tables~\ref{tab:LLM_class} \&~\ref{tab:LLM_reg} in the Appendix). The model performance is therefore robust to the type of LLM chosen. We also see the features evaluated for XGBoost models work well in Random Forest \cite{breiman2001random}, and Autogluon models (see Tables~\ref{tab:robustness_class} \&~\ref{tab:robustness_reg} in the Appendix). The features {FAMOSE} discovers appear to work well across a variety of models.

We can also test whether {FAMOSE} can be ablated to a simpler model (Appendix Tables~\ref{tab:ablation_class} \&~\ref{tab:ablation_reg}). We find, for example, that removing the ``goal'' in our prompt (``improve performance by at least 1\%'') can sometimes lead to lower performance. We believe this is because the agent decides to stop the results early without working ``hard'' to create ever better features. We similarly find, as expected, that feature selection alone is insufficient to boost model performance (thus motivating feature engineering). For example, \texttt{balance-scale} achieves ROC-AUC of 1.0 once an LLM feature related to torque is discovered, but feature selection would never discover this feature. Similarly, the model performance is reduced when we do not select features (which is needed to simplify an otherwise complex model).

In conclusion, we find: (1) {FAMOSE}'s ReAct framework can create SOTA features in both classification and regression tasks, especially for tasks  with more than 10K instances (2) creating a goal for {FAMOSE} (``improve performance by at least 1\%'') can further improve model performance, and (3) FAMOSE-based features derived from one model can improve the performance of other models.

\begin{table*}[tbh!]
\footnotesize
    \centering
    \begin{tabular}{|l|c|c|c|c|}
    \hline
Task&Baseline&OpenFE&AutoFeat&FAMOSE\\\hline
\texttt{bike} & 40.3$\pm$1.03&92.09$\pm$6.35&41.47$\pm$1.01&\textbf{40.05$\pm$0.99}\\
\texttt{crab} & 2.32$\pm$0.13&\textbf{2.26$\pm$0.15}&--&2.34$\pm$0.08\\
\texttt{cybersecurity\_attacks} & \textbf{0.82$\pm$0.0}&--&--&\textbf{0.82$\pm$0.0}\\
\texttt{housing} & 409.58$\pm$10.54&432.87$\pm$10.22&403.96$\pm$9.61&\textbf{408.56$\pm$26.34}\\
\texttt{insurance} & 192.89$\pm$13.91&190.23$\pm$11.18&\textbf{187.6$\pm$12.48}&191.64$\pm$11.16\\
\texttt{forest-fires} & 92.7$\pm$5.34&88.52$\pm$5.91&93.49$\pm$5.42&\textbf{79.49$\pm$5.87}\\
\texttt{wine-quality} & 0.64$\pm$0.02&0.75$\pm$0.03&\textbf{0.62$\pm$0.02}&0.64$\pm$0.01\\\hline
\% Reduction $\uparrow$ &0.0\%&-20.7\%&0.3\%&\textbf{2.0\%}\\
\% Tasks analyzed $\uparrow$ &\textbf{100\%} &86\%&71\%&\textbf{100\%}\\\hline
    \end{tabular}
    \caption{RMSE comparison of {FAMOSE} to SOTA feature engineering algorithms across 5 folds: OpenFE \cite{zhang2023openfe} and AutoFeat \cite{horn2019autofeat}. The ``\% Reduction'' is the mean percent reduction of RMSE across all folds and tasks. This is calculated by replacing any algorithms that could not run with the baseline performance. \textbf{Bold} indicates best model.
    }
    \label{tab:Comparison_reg}
\end{table*}

\begin{table*}[tbh!]
\footnotesize
    \centering
    \begin{tabular}{|l|c|c|c|c|c|c|}
    \hline
Task&Baseline&OpenFE&AutoFeat&FeatLLM&CAAFE&FAMOSE\\\hline
\texttt{balance-scale} & 0.914$\pm$0.01&0.999$\pm$0.001&0.977$\pm$0.025&\textbf{1.0$\pm$0.0}&0.965$\pm$0.022&\textbf{1.0$\pm$0.0}\\
\texttt{blood} & 0.698$\pm$0.03&0.681$\pm$0.048&0.677$\pm$0.033&0.426$\pm$0.063&0.697$\pm$0.03&\textbf{0.704$\pm$0.027}\\
\texttt{breast-w} & 0.989$\pm$0.007&\textbf{0.993$\pm$0.004}&\textbf{0.993$\pm$0.006}&0.994$\pm$0.004&0.99$\pm$0.01&0.989$\pm$0.008\\
\texttt{car} & 0.999$\pm$0.0&\textbf{1.0$\pm$0.0}&\textbf{1.0$\pm$0.0}&0.86$\pm$0.019&0.999$\pm$0.0&0.997$\pm$0.003\\
\texttt{cmc} & 0.704$\pm$0.019&0.693$\pm$0.013&0.683$\pm$0.019&\textbf{0.709$\pm$0.031}&0.7$\pm$0.02&0.696$\pm$0.023\\
\texttt{communities} & \textbf{0.866$\pm$0.008}&--&--&0.808$\pm$0.011&0.864$\pm$0.007&0.861$\pm$0.007\\
\texttt{credit-g} & 0.771$\pm$0.036&0.772$\pm$0.023&0.775$\pm$0.026&0.711$\pm$0.035&\textbf{0.783$\pm$0.024}&0.757$\pm$0.04\\
\texttt{eucalyptus} & 0.835$\pm$0.01&\textbf{0.907$\pm$0.009}&0.903$\pm$0.008&--&0.83$\pm$0.014&0.836$\pm$0.021\\
\texttt{heart} & 0.919$\pm$0.023&0.919$\pm$0.022&0.914$\pm$0.026&\textbf{0.92$\pm$0.021}&0.917$\pm$0.023&0.903$\pm$0.027\\
\texttt{myocardial} & \textbf{0.688$\pm$0.082}&0.686$\pm$0.093&--&0.592$\pm$0.054&0.679$\pm$0.09&0.679$\pm$0.05\\
\texttt{pc1} & 0.822$\pm$0.05&0.795$\pm$0.104&0.82$\pm$0.049&0.701$\pm$0.055&\textbf{0.834$\pm$0.033}&0.832$\pm$0.064\\
\texttt{tic-tac-toe} & 0.999$\pm$0.002&\textbf{1.0$\pm$0.0}&0.999$\pm$0.002&--&0.999$\pm$0.002&\textbf{1.0$\pm$0.001}\\
\texttt{vehicle} & 0.929$\pm$0.007&\textbf{0.952$\pm$0.007}&0.951$\pm$0.003&0.792$\pm$0.008&0.93$\pm$0.008&0.918$\pm$0.021\\\hline
Small Task \% Improvement $\uparrow$ &0.0\%&\textbf{1.04\%}&0.92\%&-7.6\%&0.47\%&0.36\%\\\hline
\texttt{adult} & \textbf{0.929$\pm$0.002}&0.896$\pm$0.002&--&0.887$\pm$0.003&\textbf{0.929$\pm$0.002}&\textbf{0.929$\pm$0.002}\\
\texttt{bank\_fraud\_base} & 0.886$\pm$0.004&0.837$\pm$0.005&--&0.718$\pm$0.007&\textbf{0.887$\pm$0.003}&0.886$\pm$0.003\\
\texttt{bank\_marketing} & \textbf{0.934$\pm$0.005}&--&--&0.738$\pm$0.02&0.933$\pm$0.004&0.928$\pm$0.013\\
\texttt{covtype} & 0.99$\pm$0.0&\textbf{0.994$\pm$0.0}&--&0.743$\pm$0.003&0.99$\pm$0.0&0.989$\pm$0.001\\
\texttt{diabetes} & \textbf{0.705$\pm$0.004}&0.673$\pm$0.003&--&--&\textbf{0.705$\pm$0.005}&\textbf{0.705$\pm$0.003}\\
\texttt{junglechess} & 0.974$\pm$0.001&\textbf{0.999$\pm$0.0}&--&0.718$\pm$0.004&0.974$\pm$0.001&0.994$\pm$0.006\\\hline
Large Task \%  Improvement $\uparrow$ &0.0\%&-1.77\%&-0.018\%&-15.9\%&0.008\%&\textbf{0.229\%}\\\hline
\% Overall Improvement $\uparrow$ &0.0\%&0.15\%&\textbf{0.62\%}&-10.2\%*&0.32\%&0.32\%\\
\% Tasks analyzed $\uparrow$ &\textbf{100\%}&89\%&58\%&84\%&\textbf{100\%}&\textbf{100\%}\\\hline
\end{tabular}
    \caption{ROC-AUC comparison of {FAMOSE} to SOTA feature engineering algorithms across 5 folds: AutoFeat \cite{horn2019autofeat}, OpenFE \cite{zhang2023openfe}, FeatLLM \cite{han2024large}, and CAAFE \cite{Hollmann2023}.  Algorithms that could not run due to errors are indicated by ``--''. The ``\% Improvement'' is the mean percent increase in ROC-AUC across folds and tasks. This is calculated by replacing any algorithms that could not run with the baseline performance. \textbf{Bold} indicates best model.}
    \label{tab:Comparison_class}
\end{table*}

\section{Discussion}
Our proposed ReAct framework, {FAMOSE} is a novel way to create SOTA model performance for classification tasks (especially large classification tasks) as well as regression tasks. Moreover, the features engineered by FAMOSE improve the performance of both weaker (e.g., Random Forest) and stronger (e.g., Autogluon) models. If FAMOSE is applied to a fast and simple model, our results suggest that an end-user could be reasonably confident that FAMOSE-engineered features can also improve the performance of a more powerful, slower, or costlier model, thereby saving time or money in the process.  

In contrast to prior methods, FAMOSE effectively simulates a data scientist by iteratively hypothesizing, testing, and refining features to improve predictions. Moreover, {FAMOSE} self-corrects code to robustly develop new features for a rich variety of datasets. We hypothesize FAMOSE's ability to create useful features is due to the LLM context window recording the features that did and did not improve model performance, which is akin to few-shot learning, allowing new features invented by an agent to improve on the mistakes of features the agent invented in previous steps. Overall, our results hint at ways that AI agents can be general problem solvers in situations that require highly inventive solutions, such as feature engineering. More specifically, with the help of post-processing checking (to reduce hallucinations), and the right tools and prompts, other agents can be able to similarly develop trustworthy, error-free outputs in highly inventive tasks.


\subsection{Limitations}
There are several limitations associated with FAMOSE. First, the ReAct framework can be expensive due to the number of tokens needed for agentic chain-of-though reasoning. In addition, {FAMOSE} appears to perform more poorly on cheaper and smaller LLMs (based on our initial exploration of {FAMOSE} with Llama 3.1-8B). Third, the benefit of {FAMOSE} depends on the background knowledge available to the LLM; for example, the LLM will perform better on a common task than a task very different from the text it was trained on. A RAG framework might help in these scenarios by inserting more bespoke background knowledge of the task to the LLM prior to the feature generation step. Finally, {FAMOSE} will need to be modified to create features for multi-label classification, but we expect such modifications would be minor.  

\section{Conclusion}
We introduced FAMOSE, a novel ReAct-based algorithm to automate feature engineering for tabular machine learning tasks. This approach reduces reliance on deep domain expertise, speeds up responses to emerging patterns, and lays the groundwork to automate the end-to-end machine learning lifecycle. {FAMOSE} utilizes agents to create features and evaluates the features again after the agent finishes its task to control for LLM hallucinations. Finally, it selects a subset of features with mRMR to optimize model performance. Using these features, we found SOTA performance improvements across classification and regression tasks, especially for large classification tasks. These features were found to work well across various models and came with explanations written by the agent's LLM, which helps humans better understand these features, in contrast to many non-LLM methods.   

\section*{Impact Statement}

This paper presents work whose goal is to advance the field of Machine Learning. There are many potential societal consequences of our work, none which we feel must be specifically highlighted here.


\newpage
\appendix
\onecolumn
\newpage
\setcounter{figure}{0}                       
\renewcommand\thefigure{S\arabic{figure}} 
\setcounter{table}{0}                       
\renewcommand\thetable{S\arabic{table}} 
\section{Appendix}
\subsection{Full LLM Prompt}
The full prompt is below\\
\noindent\fbox{\begin{minipage}{\textwidth}
\textbf{Prompt:\\}
You are a data analyst expert with full knowledge of data analysis methods.\\
    You have been given the following file:\\
    - \texttt{<file name>}, this is a CSV file useful to answer the question: \texttt{<Ex: Determine if an individual has diabetes. No diabetes, prediabetes or diabetes?>}\\
In addition, we have the following descriptions of each column: \texttt{<dict of each feature and description>}\\\\
    Your tasks are the following. Only perform one task within your code. Do not write the code for multiple tasks at once:\\\\
    - Task 1. Use all of these insights to create a large set of new features. You can use use any mathematical operations or transforms. Do not use any black box models, such as Random Forest, XGBoost, LightGBM, etc., and do not use the "Diabetes" feature in your newly generated features because that trivially improves model performance.\\\\
    - Task 2. Explain why this feature should help answer the question: \texttt{<Ex: Determine if an individual has diabetes. No diabetes, prediabetes or diabetes?>}\\
    - Task 3. Check the performance of each feature by creating a dict of the form \{"code":str(code of function  "new\_feature(df)" that when run on a dataframe returns the feature as a Pandas series)\}, saving this dict object with Pandas to\_pickle() as a file named "new\_feature\\texttt{<dataset>}\_0\_\{np.random.randint(100000)\}.pkl" and running the feature\_performance\_evaluator tool.\\\\
    - Task 4. Higher performance values are better. If the feature\_performance\_evaluator() score is over 0.01, move to the next task. Otherwise create more features and try again.\\\\
    - Task 5. Save the best performing feature use Pandas to\_pickle() to save a dict object of the form: {"code":str(code of function "new\_feature(df)" that when run on a dataframe returns the feature as a Pandas series)}; save as a file named "new\_feature\_\texttt{<dataset>}\_0\_\texttt{<random number>}.pkl", where you create a random integer with numpy. DO NOT USE pickle.dump() to save this dict object because you will get an error.
    \end{minipage}}

\subsection{Example implementation}
We show an example of how this algorithm works on one exemplar task: \texttt{balance-scale}. This task aims to predict whether a scale slides to the left, right, or is balanced, making it a 3-class classifier problem. We are given 4 features: left arm length, left arm weight, right arm length, and right arm weight. In theory, we only need one feature: the difference in torque (left arm length times left arm weight minus right arm length times right arm weight). The four features, when fed into SOTA methods gives us ROC-AUC of 0.8 for Random Forest (Table~\ref{tab:robustness_class}), or 0.91 for XGBoost (Table~\ref{tab:Comparison_class}). Interestingly, the performance is 0.995 for Autogluon, demonstrating it is a very good, although not perfect, tabular model.

FAMOSE's output is as follows, given a prompt like what is seen above:

\noindent\fbox{\begin{minipage}{\textwidth}
\textbf{Output:\\}
I'll help solve these tasks step by step.\\
\\
Thought: For Task 1, I need to create meaningful features for predicting the balance scale direction. Since this is a physics problem, I can use the principles of moments and leverage. The key insight is that the product of weight and distance gives the moment/torque on each side, and comparing these will determine which way the scale tips.\\
\\
\textless code\textgreater\\
import pandas as pd\\
import numpy as np\\
\\
\# Read the data\\
df = pd.read\_csv("../data/balance-scale\_train\_3.csv")\\
\\
\# Create features based on physics principles\\
def create\_features(df):\\
$~~~$\# Calculate moments (torque) for each side\\
$~~~$left\_moment = df['Left-Weight'] * df['Left-Distance']\\
$~~~$right\_moment = df['Right-Weight'] * df['Right-Distance']\\
$~~~$\\
$~~~$\# Calculate moment differences and ratios\\
$~~~$moment\_difference = left\_moment - right\_moment\\
$~~~$moment\_ratio = left\_moment / right\_moment\\
$~~~$\\
$~~~$\# Calculate total weights and distances\\
$~~~$total\_weight = df['Left-Weight'] + df['Right-Weight']\\
$~~~$total\_distance = df['Left-Distance'] + df['Right-Distance']\\
$~~~$\\
$~~~$\# Calculate squared terms\\
$~~~$left\_moment\_squared = left\_moment ** 2\\
$~~~$right\_moment\_squared = right\_moment ** 2\\
$~~~$\\
$~~~$\# Calculate log transformations (adding 1 to avoid log(0))\\
$~~~$log\_moment\_ratio = np.log1p(moment\_ratio)\\
$~~~$\\
$~~~$\# Create feature dictionary\\
$~~~$features = {\\
$~~~$$~~~$'moment\_difference': moment\_difference,\\
$~~~$$~~~$'moment\_ratio': moment\_ratio,\\
$~~~$$~~~$'total\_weight': total\_weight,\\
$~~~$$~~~$'total\_distance': total\_distance,\\
$~~~$$~~~$'left\_moment\_squared': left\_moment\_squared,\\
$~~~$$~~~$'right\_moment\_squared': right\_moment\_squared,\\
$~~~$$~~~$'log\_moment\_ratio': log\_moment\_ratio\\
$~~~$}\\
$~~~$\\
$~~~$return features
\end{minipage}}

\noindent\fbox{\begin{minipage}{\textwidth}
\textbf{Output (cont.):\\}
\# Create the features\\
new\_features = create\_features(df)\\
print("Created features:", list(new\_features.keys()))\\
\textless/code\textgreater\\
\\
Observation: Created features: ['moment\_difference', 'moment\_ratio', 'total\_weight', 'total\_distance', 'left\_moment\_squared', 'right\_moment\_squared', 'log\_moment\_ratio']\\
\\
Thought: For Task 2, I'll explain why these features should help predict the balance scale direction.\\
\\
\textless code\textgreater\\
explanation = """\\
The created features should help predict the balance scale direction for the following reasons:\\
\\
1. moment\_difference (left\_moment - right\_moment):\\
   - If positive, the scale tips left\\
   - If negative, the scale tips right\\
   - If zero, the scale is balanced\\
   This is directly based on the physics principle of torque.\\
\\
2. moment\_ratio (left\_moment / right\_moment):\\
   - If \textgreater 1, scale tips left\\
   - If \textless 1, scale tips right\\
   - If = 1, scale is balanced\\
   This gives a relative measure of the imbalance.\\
\\
3. total\_weight and total\_distance:\\
   These provide context about the overall scale of the problem.\\
\\
4. squared moments:\\
   Captures non-linear relationships and emphasizes larger differences.\\
\\
5. log\_moment\_ratio:\\
   Transforms the ratio to handle skewed distributions better.\\
"""\\
\\
print(explanation)\\
\textless/code\textgreater
\end{minipage}}

\noindent\fbox{\begin{minipage}{\textwidth}
\textbf{Output (cont.):\\}
Observation: Printed explanation of feature importance and physics principles.\\
\\
Thought: For Task 3, I'll test the performance of one of our most promising features - the moment\_difference, as it directly captures the physical principle determining the scale's direction.\\
\\
\textless code\textgreater\\
import pickle\\
\\
\# Create the feature function for moment\_difference\\
feature\_code = """\\
def new\_feature(df):\\
$~~~$return df['Left-Weight'] * df['Left-Distance'] - df['Right-Weight'] * df['Right-Distance']\\
"""\\
\\
\# Create dictionary with the feature code\\
feature\_dict = {"code": feature\_code}\\
\\
\# Save the feature dictionary using pandas\\
random\_number = np.random.randint(100000)\\
filename = f"new\_feature\_balance-scale\_0\_{random\_number}.pkl"\\
pd.to\_pickle(feature\_dict, filename)\\
\\
\# Evaluate the feature performance\\
performance = feature\_performance\_evaluator(filename)\\
print(f"Feature performance score: {performance}")\\
\textless/code\textgreater\\
\\
Observation: Feature performance score: 0.815\\
\\
Thought: For Task 4, since our feature\_performance\_evaluator score (0.815) is well above 0.01, we can proceed to Task 5.\\
\\
\textless code\textgreater\\
\# Create the final feature dictionary with our best performing feature\\
final\_feature\_code = """\\
def new\_feature(df):\\
$~~~$return df['Left-Weight'] * df['Left-Distance'] - df['Right-Weight'] * df['Right-Distance']\\
"""\\
\\
final\_feature\_dict = {"code": final\_feature\_code}\\
\\
\# Save with pandas to\_pickle using a random number\\
random\_number = np.random.randint(100000)\\
final\_filename = f"new\_feature\_balance-scale\_0\_{random\_number}.pkl"\\
pd.to\_pickle(final\_feature\_dict, final\_filename)\\
\\
final\_answer(f"Successfully completed all tasks. Final feature saved as {final\_filename} with performance score of 0.815")\\
\textless/code\textgreater
\end{minipage}}

After this output, it prints the following observation:

\noindent\fbox{\begin{minipage}{\textwidth}
\textbf{Observations:}
Created features: ['moment\_difference', 'moment\_ratio', 'total\_weight', 'total\_distance', 'left\_moment\_squared', 'right\_moment\_squared', 'log\_moment\_ratio']\\
\\
The created features should help predict the balance scale direction for the following reasons:\\
\\
1. moment\_difference (left\_moment - right\_moment):\\
   - If positive, the scale tips left\\
   - If negative, the scale tips right\\
   - If zero, the scale is balanced\\
   This is directly based on the physics principle of torque.\\
\\
2. moment\_ratio (left\_moment / right\_moment):\\
   - If $$>$$ 1, scale tips left\\
   - If $$<$$ 1, scale tips right\\
   - If = 1, scale is balanced\\
   This gives a relative measure of the imbalance.\\
\\
3. total\_weight and total\_distance:\\
   These provide context about the overall scale of the problem.\\
\\
4. squared moments:\\
   Captures non-linear relationships and emphasizes larger differences.\\
\\
5. log\_moment\_ratio:\\
   Transforms the ratio to handle skewed distributions better.\\
\\
Feature performance score: 0.09026\\
Last output from code snippet:\\
Successfully completed all tasks. Final feature saved as new\_feature\_balance-scale\_0\_33885.pkl with performance score of 0.815
\end{minipage}}

We notice that the LLM hallucinates with a statement ``performance score of 0.815'', hence why all observations are checked post-facto (where we confirm the true ROC-AUC performance improvement of 0.09026 within a validation dataset, separated from the test dataset. This represents a change in ROC-AUC from 0.91 to 1.0 within that dataset. We then use mRMR to optimize the number of features needed, which turns out to exactly the above feature. The final model is simplified to become a single feature model with perfect prediction. For context, other methods, even with comparable accuracy, will often have more features.

\begin{table}[tbh!]
\footnotesize
    \centering
    \begin{tabular}{|l|l|c|c|c|}
\hline
    Task & \# Datapoints & \# Features & Imbalance & Target Variance \\\hline

adult & 48842 & 15 & 0.24 & -- \\
arrhythmia & 452 & 280 & 0.0 & -- \\
balance-scale & 625 & 5 & 0.08 & -- \\
bank\_fraud\_base & 1000000 & 32 & 0.01 & -- \\
bank\_marketing & 45211 & 17 & 0.12 & -- \\
blood & 748 & 5 & 0.24 & -- \\
breast-w & 699 & 10 & 0.34 & -- \\
communities & 1994 & 103 & 0.33 & -- \\
credit-g & 1000 & 21 & 0.3 & -- \\
diabetes & 253680 & 22 & 0.02 & -- \\
car & 1728 & 7 & 0.04 & -- \\
covtype & 581012 & 55 & 0.0 & -- \\ 
cmc & 1473 & 10 & 0.23 & -- \\
eucalyptus & 736 & 20 & 0.14 & -- \\
heart & 918 & 12 & 0.45 & -- \\
junglechess & 44819 & 7 & 0.1 & -- \\
myocardial & 686 & 92 & 0.22 & -- \\
pc1 & 1109 & 22 & 0.07 & -- \\
tic-tac-toe & 958 & 10 & 0.35 & -- \\
vehicle & 846 & 19 & 0.24 & -- \\
\hline
crab & 3893 & 9 & -- & 3.2 \\
bike & 17379 & 13 & -- & 179.7 \\
forest-fires & 517 & 13 & -- & 80.2 \\
housing & 20640 & 10 & -- & 1024.5 \\
insurance & 1338 & 7 & -- & 386.2 \\
wine & 6497 & 13 & -- & 0.9 \\
cybersecurity\_attacks & 40000 & 25 & -- & 0.8 \\\hline
    \end{tabular}
    \caption{Data statistics: number of datapoints, number of features, data imbalance (if a classification task) or target variance (if a regression task).}
    \label{tab:data_stats}
\end{table}

\begin{table*}[tbh!]
\footnotesize
\centering
\setlength{\tabcolsep}{6pt}
\renewcommand{\arraystretch}{1.15}
\begin{tabular}{|p{2cm}|p{6cm}|p{6.0cm}|}
\hline
Task Type & Task & Prediction Task \\\hline
\multirow{22}{*}{\textbf{Classification}}
& \texttt{adult}~\cite{adult1996} & Earn more than \$50{,}000 per year (yes/no) \\
& \texttt{arrhythmia}~\cite{arrhythmia_5} & Presence of cardiac arrhythmia (16 classes) \\
& \texttt{balance-scale}~\cite{balance_scale_12} & Balance direction: left, right, or balanced \\
& \texttt{breast-w}~\cite{breast_cancer_wisconsin} & Breast mass malignant (M) or benign (B) \\
& \texttt{blood}~\cite{blood_transfusion_service_center_176} & Blood donation occurred (yes/no) \\
& \texttt{car}~\cite{car_evaluation_19} & Car acceptability rating (4 classes) \\
& \texttt{diabetes}~\cite{diabetes} & Diabetes status: none, prediabetes, diabetes \\
& \texttt{cmc}~\cite{contraceptive_method_choice_30} & Contraceptive method choice (3 classes) \\
& \texttt{communities}~\cite{communities_and_crime_183} & Violent crime rate: low, medium, or high \\
& \texttt{covtype}~\cite{covertype_31} & Forest cover type (7 classes) \\
& \texttt{credit-g}~\cite{creditg} & Credit approved (yes/no) \\
& \texttt{eucalyptus}~\cite{eucalyptus} & Seedlot utility classification \\
& \texttt{heart}~\cite{heart} & Heart disease present (yes/no) \\
& \texttt{junglechess}~\cite{jungle} & Winner of Jungle Chess endgame \\
& \texttt{myocardial}~\cite{myocardial_infarction_complications_579} & Chronic heart failure present (yes/no) \\
& \texttt{pc1}~\cite{pc1} & Software defect present (1/0) \\
& \texttt{tic-tac-toe}~\cite{tic-tac-toe} & First player wins (positive/negative) \\
& \texttt{vehicle}~\cite{vehicle} & Car acceptability rating (4 classes) \\
& \texttt{bank\_fraud\_base}~\cite{bank_fraud} & Fraudulent applicant (yes/no) \\
& \texttt{travel}~\cite{booking} & Hotel booking vs.\ link click \\
& \texttt{bank\_marketing}~\cite{bank_marketing} & Subscribed to term deposit (yes/no) \\\hline
\multirow{8}{*}{\textbf{Regression}}
& \texttt{bike}~\cite{bike} & Total rental bike count \\
& \texttt{crab}~\cite{crab_age} & Crab age estimation \\
& \texttt{housing}~\cite{housing} & House price estimation \\
& \texttt{insurance}~\cite{insurance} & Individual medical cost \\
& \texttt{forest-fires}~\cite{forestfire} & Burned forest area (Portugal) \\
& \texttt{wine-quality}~\cite{wine} & Wine quality score (0--10) \\
& \texttt{cybersecurity\_attacks}~\cite{cyber_attacks} & Alert severity level \\\hline
\end{tabular}
\caption{Tasks used to evaluate {FAMOSE}.}
\label{tab:data_desc}
\end{table*}

\begin{table}[tbh!]
\footnotesize
    \centering
    \begin{tabular}{|p{3cm}|p{7cm}|}
\hline
age& ``the age of an individual"\\\hline
workclass& ``a general term to represent the employment status of an individual"\\\hline
fnlwgt& ``the number of units in the target population that the responding unit represents"\\\hline
education& ``the highest level of education achieved by an individual"\\\hline
educational-num	& ``the highest level of education achieved in numerical form"\\\hline
marital-status& ``marital status of an individual"\\\hline
occupation& ``the general type of occupation of an individual"\\\hline
relationship& ``what this individual is relative to others"\\\hline
race	& ``race"\\\hline
gender	& ``gender"\\\hline
capital-gain& ``capital gain last year"\\\hline
capital-loss	& ``capital loss last year"\\\hline
hours-per-week& ``the hours an individual has reported to work per week"\\\hline
native-country	& ``country of origin for an individual"\\\hline
    \end{tabular}
    \caption{Example feature descriptions for the \texttt{adult} task.}
    \label{tab:feature_desc1}
\end{table}

\begin{table}[tbh!]
\footnotesize
    \centering
    \begin{tabular}{|p{3cm}|p{7cm}|}
	\hline
age& ``age in years"\\\hline
job	& ``type of job (categorical: 'admin.', 'unknown', 'unemployed', 'management', 'housemaid', 'entrepreneur', 'student', 'blue-collar', 'self-employed,retired', 'technician', 'services')"\\\hline
marital& ``marital status (categorical: 'married','divorced','single'; note: 'divorced' means divorced or widowed)"\\\hline
education	& ``(categorical: 'unknown','secondary','primary','tertiary')"\\\hline
default	& ``has credit in default? (binary: 'yes','no')"\\\hline
balance	& ``average yearly balance, in euros (numeric)"\\\hline
housing& ``has housing loan? (binary: 'yes','no')"\\\hline
loan	& ``has personal loan? (binary: 'yes','no')"\\\hline
contact& ``contact communication type (categorical: 'unknown','telephone','cellular')"\\\hline
day	& ``last contact day of the month (numeric)"\\\hline
month	& ``last contact month of year (categorical: 'jan', 'feb', 'mar', ..., 'nov', 'dec')"\\\hline
duration& ``last contact duration, in seconds (numeric)"\\\hline
campaign& ``number of contacts performed during this campaign and for this client (numeric, includes last contact)"\\\hline
pdays	& ``number of days that passed by after the client was last contacted from a previous campaign (numeric, -1 means client was not previously contacted)"\\\hline
previous	& ``number of contacts performed before this campaign and for this client (numeric)"\\\hline
poutcome& ``outcome of the previous marketing campaign (categorical: 'unknown','other','failure','success')"\\\hline
    \end{tabular}
    \caption{Example feature descriptions for the \texttt{bank-marketing} task.}
    \label{tab:feature_desc2}
\end{table}

\begin{figure}[tbh!]
    \centering
    \includegraphics[width=1\linewidth]{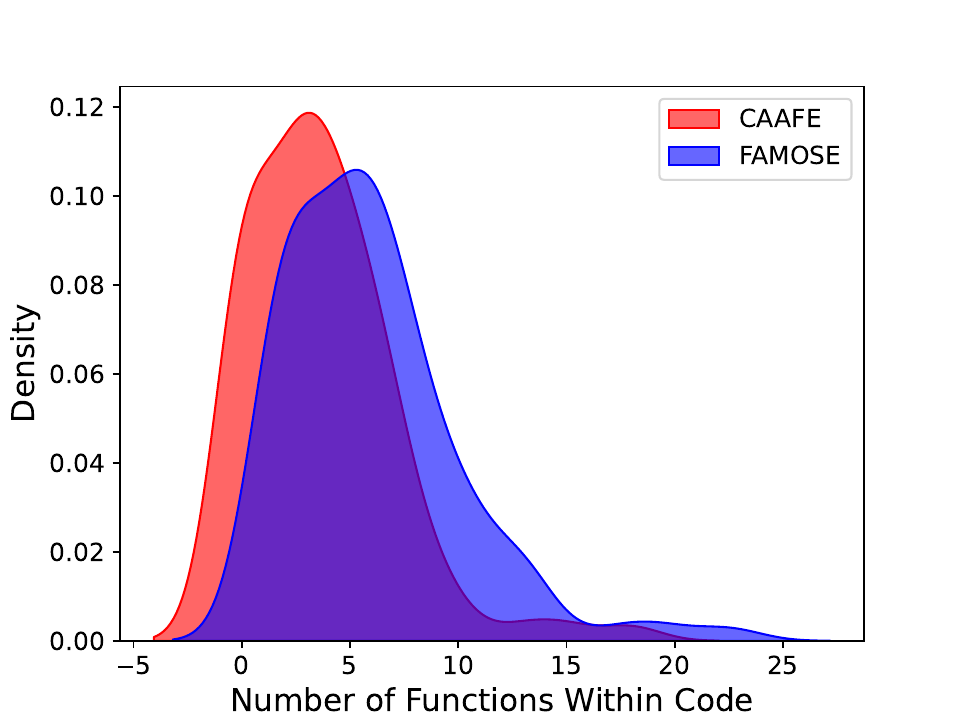}
    \caption{Number of functions used in CAAFE and {FAMOSE} code across all task folds (5$\times$ number of tasks).}
    \label{fig:FunctDist}
\end{figure}

\begin{figure}[tbh!]
    \centering
    \includegraphics[width=1\linewidth]{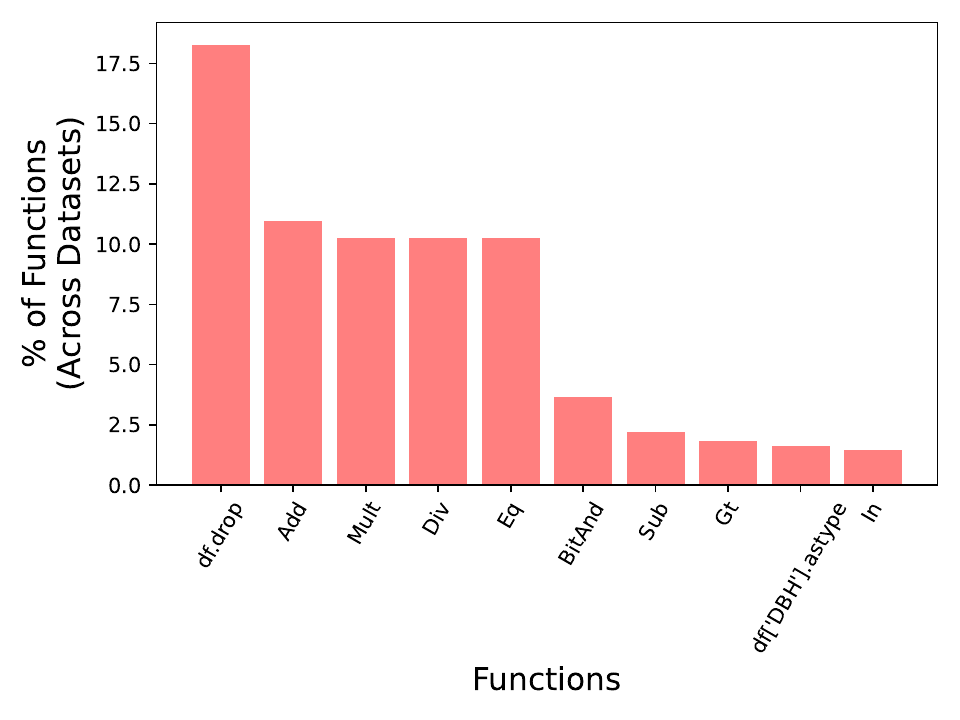}
    \caption{Frequency each function used in {CAAFE} code across all task folds (5$\times$ number of tasks).}
    \label{fig:FunctFreqCAAFE}
\end{figure}

\begin{figure}[tbh!]
    \centering
    \includegraphics[width=1\linewidth]{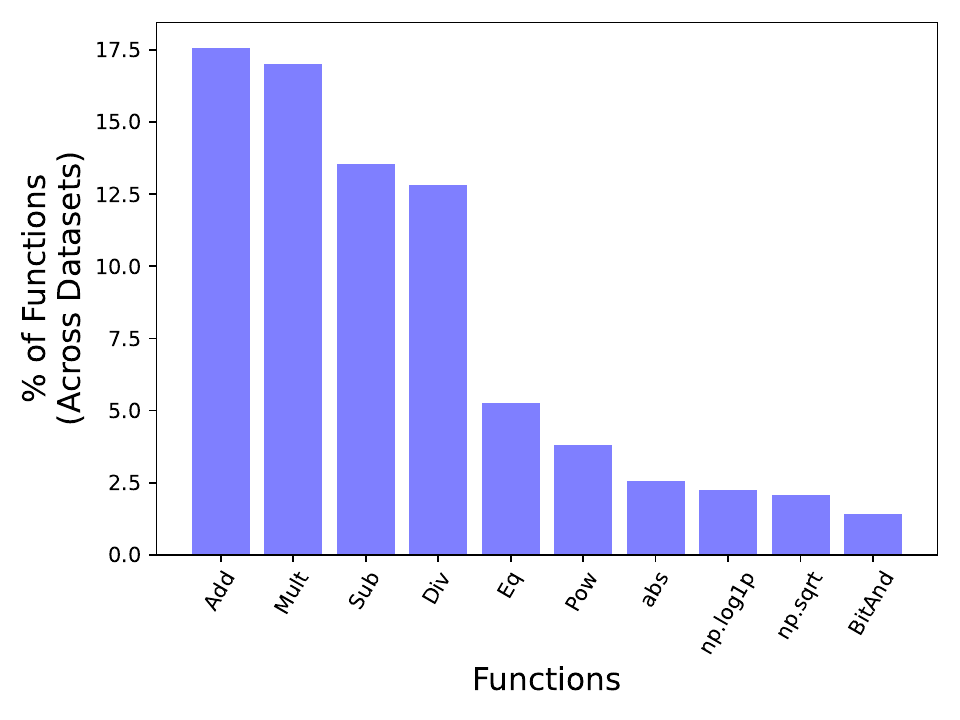}
    \caption{Frequency each function used in {FAMOSE} code across all task folds (5$\times$ number of tasks).}
    \label{fig:FunctFreqFAMOSE}
\end{figure}

\begin{figure}[tbh!]
    \centering
    \includegraphics[width=1\linewidth]{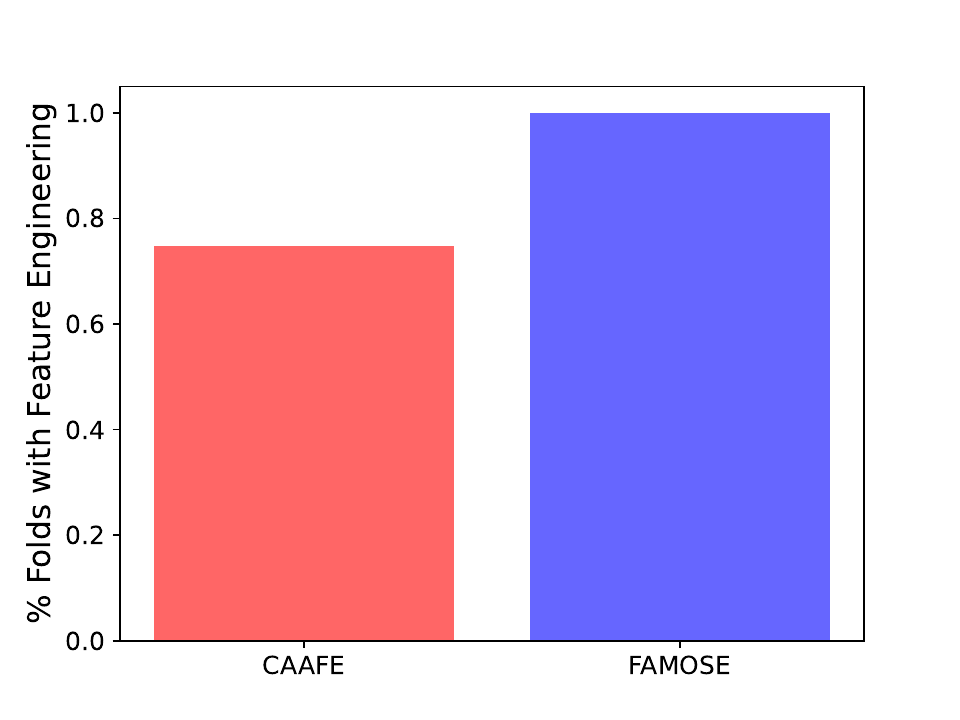}
    \caption{Percentage of task folds (5$\times$ number of tasks) where {FAMOSE} or CAAFE features were added.}
    \label{fig:PercentFE}
\end{figure}

\begin{table*}[tbh!]
\footnotesize
    \centering
    \begin{tabular}{|l|c|c|c|}
    \hline
&Baseline&FAMOSE (Claude) &FAMOSE (Deepseek)\\\hline
\texttt{adult} & \textbf{0.929$\pm$0.002}&\textbf{0.929$\pm$0.002}&\textbf{0.929$\pm$0.002}\\
\texttt{balance-scale} & 0.914$\pm$0.01&\textbf{1.0$\pm$0.0}&\textbf{1.0$\pm$0.0}\\
\texttt{bank\_fraud\_base} & \textbf{0.886$\pm$0.004}&\textbf{0.886$\pm$0.003}&0.885$\pm$0.004\\
\texttt{bank\_marketing} &\textbf{ 0.934$\pm$0.005}&0.928$\pm$0.013&0.933$\pm$0.004\\
\texttt{blood} & 0.698$\pm$0.03&\textbf{0.704$\pm$0.027}&0.687$\pm$0.009\\
\texttt{breast-w} & \textbf{0.989$\pm$0.007}&\textbf{0.989$\pm$0.008}&0.986$\pm$0.008\\
\texttt{car} & \textbf{0.999$\pm$0.0}&0.997$\pm$0.003&\textbf{0.999$\pm$0.002}\\
\texttt{communities} & \textbf{0.866$\pm$0.008}&0.861$\pm$0.007&0.864$\pm$0.005\\
\texttt{covtype} & 0.990$\pm$0.0&0.989$\pm$0.001&\textbf{0.992$\pm$0.001}\\
\texttt{cmc} & \textbf{0.704$\pm$0.019}&0.696$\pm$0.023&0.698$\pm$0.026\\
\texttt{credit-g} & \textbf{0.771$\pm$0.036}&0.757$\pm$0.04&0.748$\pm$0.022\\
\texttt{diabetes} & \textbf{0.705$\pm$0.004}&\textbf{0.705$\pm$0.003}&\textbf{0.705$\pm$0.003}\\
\texttt{eucalyptus} & 0.835$\pm$0.01&\textbf{0.836$\pm$0.021}&0.835$\pm$0.03\\
\texttt{heart} & \textbf{0.919$\pm$0.023}&0.903$\pm$0.027&0.916$\pm$0.021\\
\texttt{junglechess} & 0.974$\pm$0.001&0.994$\pm$0.006&\textbf{0.997$\pm$0.001}\\
\texttt{pc1} & 0.822$\pm$0.05&\textbf{0.832$\pm$0.064}&\textbf{0.832$\pm$0.068}\\
\texttt{myocardial} & \textbf{0.688$\pm$0.082}&0.679$\pm$0.05&0.676$\pm$0.09\\
\texttt{tic-tac-toe} & 0.999$\pm$0.002&\textbf{1.0$\pm$0.001}&0.997$\pm$0.004\\
\texttt{vehicle} & 0.929$\pm$0.007&0.918$\pm$0.021&\textbf{0.930$\pm$0.009}\\\hline
\% Improvement $\uparrow$ &0.0\%&\textbf{0.32\%}&0.29\%\\\hline
    \end{tabular}
    \caption{FAMOSE classifier performance (ROC-AUC) for Claude and Deepseek. We notice that the algorithm works well for both closed-source (Claude 3.5 Sonnet V2) and open-source (Deepseek-R1) LLMs.}
    \label{tab:LLM_class}
\end{table*}

\begin{table*}[tbh!]
\footnotesize
    \centering
    \begin{tabular}{|l|c|c|c|}
    \hline
&Baseline&FAMOSE (Sonnet 3.5 V2)&FAMOSE (Deepseek)\\\hline
\texttt{bike} & 40.3$\pm$1.03&40.05$\pm$0.99&\textbf{39.79$\pm$0.73}\\
\texttt{crab} & 2.32$\pm$0.13&2.34$\pm$0.08&\textbf{2.31$\pm$0.16}\\
\texttt{housing} & 409.58$\pm$10.54&408.56$\pm$26.34&\textbf{396.44$\pm$19.22}\\
\texttt{insurance} & 192.89$\pm$13.91&191.64$\pm$11.16&\textbf{190.16$\pm$11.96}\\
\texttt{forest-fires} & 92.7$\pm$5.34&79.49$\pm$5.87&\textbf{79.47$\pm$4.6}\\
\texttt{wine-quality} & \textbf{0.64$\pm$0.02}&\textbf{0.64$\pm$0.01}&\textbf{0.64$\pm$0.02}\\
\texttt{cybersecurity\_attacks} & \textbf{0.82$\pm$0.0}&\textbf{0.82$\pm$0.0}&\textbf{0.82$\pm$0.0}\\\hline
\% Reduction $\uparrow$ &0.0\%&2.0\%&\textbf{2.8\%}\\\hline
    \end{tabular}
    \caption{FAMOSE regression performance (RMSE) for Claude and Deepseek. Bold are best-performing models. We notice that the algorithm works well for both closed-source (Claude 3.5 Sonnet V2) and open-source (Deepseek-R1) LLMs, with a slight edge for Deepseek-R1. }
    \label{tab:LLM_reg}
\end{table*}

\begin{table*}[tbh!]
\footnotesize
    \centering
    \begin{tabular}{|l|c|c|c|c|c|}
    \hline
&    \multicolumn{2}{|c|}{Autogluon} & \multicolumn{2}{|c|}{Random Forest}\\\hline
&Baseline & {FAMOSE} &Baseline&FAMOSE\\\hline
\texttt{adult} & \textbf{0.929$\pm$0.003}&\textbf{0.929$\pm$0.002}&\textbf{0.903$\pm$0.002}&0.901$\pm$0.005\\
\texttt{balance-scale} & 0.995$\pm$0.004&\textbf{1.0$\pm$0.0}&0.806$\pm$0.009&\textbf{1.0$\pm$0.0}\\
\texttt{bank\_marketing} & \textbf{0.938$\pm$0.004}&0.937$\pm$0.006&\textbf{0.929$\pm$0.004}&0.928$\pm$0.003\\
\texttt{bank\_fraud\_base} & 0.877$\pm$0.01&\textbf{0.884$\pm$0.006}&0.834$\pm$0.003&\textbf{0.835$\pm$0.004}\\
\texttt{blood} & \textbf{0.757$\pm$0.035}&0.747$\pm$0.029&0.686$\pm$0.042&\textbf{0.706$\pm$0.03}\\
\texttt{breast-w} &0.993$\pm$0.005&\textbf{0.994$\pm$0.005}&0.991$\pm$0.006&\textbf{0.992$\pm$0.006}\\
\texttt{car} & \textbf{1.0$\pm$0.0}&0.998$\pm$0.004&\textbf{0.994$\pm$0.002}& 0.992$\pm$0.003\\
\texttt{communities} & \textbf{0.866$\pm$0.008}&0.862$\pm$0.009&\textbf{0.869$\pm$0.007}&0.863$\pm$0.008\\
\texttt{cmc} & \textbf{0.726$\pm$0.014}&0.708$\pm$0.019&\textbf{0.698$\pm$0.014}&0.688$\pm$0.024\\
\texttt{covtype} & \textbf{0.998$\pm$0.0}&\textbf{0.998$\pm$0.0}&\textbf{0.997$\pm$0.0}&\textbf{0.997$\pm$0.001}\\
\texttt{credit-g} & 0.75$\pm$0.042&\textbf{0.761$\pm$0.016}&\textbf{0.791$\pm$0.022}&0.764$\pm$0.026\\
\texttt{diabetes} & 0.699$\pm$0.005&\textbf{0.7$\pm$0.004}&\textbf{0.676$\pm$0.002}&0.675$\pm$0.002\\
\texttt{eucalyptus} & 0.879$\pm$0.028&\textbf{0.889$\pm$0.021}&\textbf{0.886$\pm$0.012}&0.882$\pm$0.015\\
\texttt{heart} & \textbf{0.928$\pm$0.023}&0.92$\pm$0.027&\textbf{0.927$\pm$0.028}&0.91$\pm$0.037\\
\texttt{junglechess} & 0.998$\pm$0.001&\textbf{1.0$\pm$0.0}&0.932$\pm$0.001&\textbf{0.996$\pm$0.002}\\
\texttt{myocardial} & 0.69$\pm$0.075&\textbf{0.699$\pm$0.051}&\textbf{0.709$\pm$0.087}&0.676$\pm$0.041\\
\texttt{pc1} & \textbf{0.824$\pm$0.067}&0.82$\pm$0.079&0.84$\pm$0.044&\textbf{0.855$\pm$0.053}\\
\texttt{tic-tac-toe} & \textbf{0.999$\pm$0.001}&\textbf{0.999$\pm$0.002}&0.998$\pm$0.004&\textbf{0.999$\pm$0.002}\\
\texttt{vehicle} & \textbf{0.938$\pm$0.014}&0.933$\pm$0.021&\textbf{0.93$\pm$0.009}&0.923$\pm$0.019
\\\hline    
\% Improvement $\uparrow$ & 0.0\%&\textbf{0.02\%}&0.0\%&\textbf{1.2\%}\\\hline
\end{tabular}
    \caption{Robustness of classifier performance (ROC-AUC) for {FAMOSE} XGBoost features applied to different models (Random forest and Autogluon). \textbf{Bold} indicates the best model.}
    \label{tab:robustness_class}
\end{table*}

\begin{table*}[tbh!]
\footnotesize
    \centering
    \begin{tabular}{|l|c|c|c|c|c|}
    \hline
&    \multicolumn{2}{|c|}{Autogluon} & \multicolumn{2}{|c|}{Random Forest}\\\hline
&Baseline & {FAMOSE} &Baseline&FAMOSE\\\hline
\texttt{bike} & \textbf{35.37$\pm$0.92}&35.41$\pm$0.96&42.02$\pm$1.53&\textbf{41.97$\pm$1.23}\\
\texttt{crab} & 2.13$\pm$0.11&\textbf{2.1$\pm$0.13}&2.19$\pm$0.09&\textbf{2.18$\pm$0.09}\\
\texttt{housing} & 384.31$\pm$15.11&\textbf{381.08$\pm$18.47}&418.38$\pm$11.23&\textbf{416.28$\pm$14.05}\\
\texttt{insurance} &\textbf{166.59$\pm$13.31}&174.56$\pm$16.7&\textbf{178.82$\pm$8.87}&183.44$\pm$11.0\\
\texttt{forest-fires} &80.89$\pm$6.11&\textbf{79.16$\pm$5.59}&83.13$\pm$5.04&\textbf{79.22$\pm$5.68}\\
\texttt{wine-quality} &\textbf{0.61$\pm$0.01}&\textbf{0.61$\pm$0.01}&\textbf{0.61$\pm$0.01}&\textbf{0.61$\pm$0.02}\\
\texttt{cybersecurity\_attacks} & \textbf{0.82$\pm$0.0}&\textbf{0.82$\pm$0.0}&0.83$\pm$0.0&\textbf{0.82$\pm$0.0}\\
\hline
\% Reduction $\uparrow$ &\textbf{0.0\%}&-0.1\%&0.0\%&\textbf{0.5\%}\\\hline
\end{tabular}
    \caption{Robustness of classifier performance (RMSE) for {FAMOSE} XGBoost features applied to different models (Random forest and Autogluon). \textbf{Bold} indicates the best model.}
    \label{tab:robustness_reg}
\end{table*}

\begin{table*}[tbh!]
\footnotesize
    \centering
    \begin{tabular}{|l|c|c|c|}
    \hline
&No goal in prompt&Only feature selection&No feature selection\\\hline
\texttt{bike} & 40.19$\pm$0.89&40.3$\pm$1.03&40.43$\pm$1.54\\
\texttt{crab} & 2.34$\pm$0.14&2.34$\pm$0.12&2.34$\pm$0.09\\
\texttt{cybersecurity\_attacks} & 0.82$\pm$0.0&0.82$\pm$0.0&0.83$\pm$0.0\\
\texttt{forest-fires} & 79.87$\pm$4.98&79.54$\pm$5.89&92.93$\pm$5.24\\
\texttt{housing} & 409.03$\pm$26.18&409.58$\pm$10.54&418.51$\pm$38.84\\
\texttt{insurance} & 186.53$\pm$12.52&184.43$\pm$16.86&193.95$\pm$13.82\\
\texttt{wine-quality} & 0.64$\pm$0.01&0.64$\pm$0.01&0.64$\pm$0.01\\\hline
\% Reduction $\uparrow$ &2.2\%&2.4\%&-0.8\%\\\hline
    \end{tabular}
    \caption{Ablation for regression tasks. We see a slight improvement in the RMSE reduction when we only select features, although the difference is not significant (Wilcoxon signed rank test p-value$=0.3$), but the difference between {FAMOSE} and no feature selection is significant (Wilcoxon signed rank test p-value$=0.007$). Overall, our results show FAMOSE performs best when we have a goal and feature selection.}
    \label{tab:ablation_reg}
\end{table*}

\begin{table*}[tbh!]
\footnotesize
    \centering
    \begin{tabular}{|l|c|c|c|}
    \hline
&No goal in prompt&Only feature selection&No feature selection\\\hline
\texttt{adult} & 0.928$\pm$0.002&0.929$\pm$0.002&0.929$\pm$0.002\\
\texttt{balance-scale} & 1.0$\pm$0.0&0.914$\pm$0.01&1.0$\pm$0.0\\
\texttt{bank\_fraud\_base} & 0.885$\pm$0.004&0.885$\pm$0.005&0.881$\pm$0.01\\
\texttt{bank\_marketing} & 0.928$\pm$0.005&0.933$\pm$0.004&0.929$\pm$0.011\\
\texttt{blood} & 0.678$\pm$0.038&0.688$\pm$0.024&0.679$\pm$0.024\\
\texttt{breast-w} & 0.99$\pm$0.008&0.989$\pm$0.006&0.985$\pm$0.011\\
\texttt{car} & 0.999$\pm$0.001&0.997$\pm$0.002&0.998$\pm$0.002\\
\texttt{cmc} & 0.691$\pm$0.01&0.704$\pm$0.023&0.7$\pm$0.015\\
\texttt{communities} & 0.859$\pm$0.005&0.861$\pm$0.007&0.866$\pm$0.007\\
\texttt{covtype} & 0.992$\pm$0.0&0.989$\pm$0.0&0.989$\pm$0.002\\
\texttt{credit-g} & 0.755$\pm$0.021&0.756$\pm$0.036&0.758$\pm$0.04\\
\texttt{diabetes} & 0.705$\pm$0.004&0.705$\pm$0.003&0.706$\pm$0.004\\
\texttt{eucalyptus} & 0.831$\pm$0.024&0.836$\pm$0.027&0.825$\pm$0.023\\
\texttt{heart} & 0.913$\pm$0.02&0.916$\pm$0.019&0.914$\pm$0.023\\
\texttt{junglechess} & 0.999$\pm$0.001&0.974$\pm$0.001&0.994$\pm$0.006\\
\texttt{myocardial} & 0.664$\pm$0.077&0.663$\pm$0.067&0.677$\pm$0.076\\
\texttt{pc1} & 0.826$\pm$0.067&0.825$\pm$0.056&0.82$\pm$0.054\\
\texttt{tic-tac-toe} & 0.996$\pm$0.006&1.0$\pm$0.001&0.998$\pm$0.003\\
\texttt{vehicle} & 0.929$\pm$0.004&0.931$\pm$0.01&0.924$\pm$0.007\\\hline
\% Improvement $\uparrow$ &0.0\%&-0.38\%&0.04\%\\\hline
    \end{tabular}
    \caption{Ablation for classifier tasks. Due to the small changes in ROC-AUC, the results are not statistically significant, except for only feature selection, which is significantly worse than the baseline and {FAMOSE} (Wilcoxon signed rank test p-value$<0.04$), which suggests that we need to go beyond feature selection along to achieve high performance.  Overall, our results show FAMOSE performs best when we have a goal and feature selection.}
    \label{tab:ablation_class}
\end{table*}


\end{document}